\documentclass[10pt,journal]{IEEEtran}

\usepackage{amsmath}
\usepackage{graphicx}
\usepackage{cite}
\usepackage{times}
\usepackage{epsfig}
\usepackage{graphicx}
\usepackage{amsmath}
\usepackage{amssymb}
\usepackage{booktabs}
\usepackage{multirow}
\usepackage[pagebackref=true,breaklinks=true,letterpaper=true,colorlinks,bookmarks=false]{hyperref}

\ifCLASSINFOpdf
\else
\fi
\makeatletter
\newcommand*{\rom}[1]{\expandafter\@slowromancap\romannumeral #1@}
\makeatother

\begin{document}
%
\author{Abrar~H.~Abdulnabi,~\IEEEmembership{Student~Member,~IEEE},
	 Gang~Wang,~\IEEEmembership{Member,~IEEE}, , Jiwen~Lu,~\IEEEmembership{Member,~IEEE}
	and~Kui~Jia,~\IEEEmembership{Member,~IEEE}
	\thanks{A. H. Abdulnabi is working with both the Rapid-Rich Object Search (ROSE) Lab at the Nanyang Technological University, Singapore, and the Advanced Digital Sciences Center (ADSC), Illinois at Singapore Pt Ltd, Singapore, Email: abrarham001@ntu.edu.sg. G. Wang is with the Department of Electrical and Electronic Engineering, Nanyang Technological University, Singapore, Email: wanggang@ntu.edu.sg. J. Lu is with the Department of Automation, Tsinghua University, Beijing, 100084, China, Email: elujiwen@gmail.com. K. Jia is with the Department of Computer and Information Science, Faculty of Science and Technology, University of Macau, Macau SAR, China, Email: kuijia@umac.mo. Address of ADSC: Advanced Digital Sciences Center, 1 Fusionopolis Way, Illinois at Singapore, Singapore 138632. Address of ROSE: The Rapid-Rich Object Search Lab, School of Electrical and Electronic Engineering, Nanyang Technological University, Singapore, 637553.}
}

\title{Multi-task CNN Model for Attribute Prediction}
\maketitle

\begin{abstract}
This paper proposes a joint multi-task learning algorithm to better predict attributes in images using deep convolutional neural networks (CNN). We consider learning binary semantic attributes through a multi-task CNN model, where each CNN will predict one binary attribute. The multi-task learning allows CNN models to simultaneously share visual knowledge among different attribute categories. Each CNN will generate attribute-specific feature representations, and then we apply multi-task learning on the features to predict their attributes. In our multi-task framework, we propose a method to decompose the overall model's parameters into a latent task matrix and combination matrix. Furthermore, under-sampled classifiers can leverage shared statistics from other classifiers to improve their performance. Natural grouping of attributes is applied such that attributes in the same group are encouraged to share more knowledge. Meanwhile, attributes in different groups will generally compete with each other, and consequently share less knowledge. We show the effectiveness of our method on two popular attribute datasets.
\end{abstract}

\begin{IEEEkeywords}
Semantic Attributes, Multi-task learning, Deep CNN, Latent tasks matrix.
\end{IEEEkeywords}

%
\IEEEpeerreviewmaketitle

\section{Introduction}
%
%
%
\IEEEPARstart{U}{sing} semantic properties, or attributes, to describe objects is a technique that has attracted much attention in visual recognition research \cite{Farhadi09describingobjects, Lampert09learningto}. This is due to the fact that learning an object's attributes provides useful and detailed knowledge about it, and also serves as a bridge between low-level features and high-level categories. Various multimedia applications can benefit from attributes, among which are the following: knowledge transfer, information sharing between different target tasks, multimedia content analysis and recommendation, multimedia search and retrieval \cite{Siddiquie5995329, Farhadi09describingobjects, Lampert09learningto, JayaramanG14, Ferrari07, Russakovsky_attributelearning, LEGO3-contextContent, LEGO4-multiLabel, LEGO5-SemanticDictance,LEGO1-structureModel}.\\

\begin{figure}
	\centering

	\begin{center}
		\includegraphics[width=0.35\textwidth]{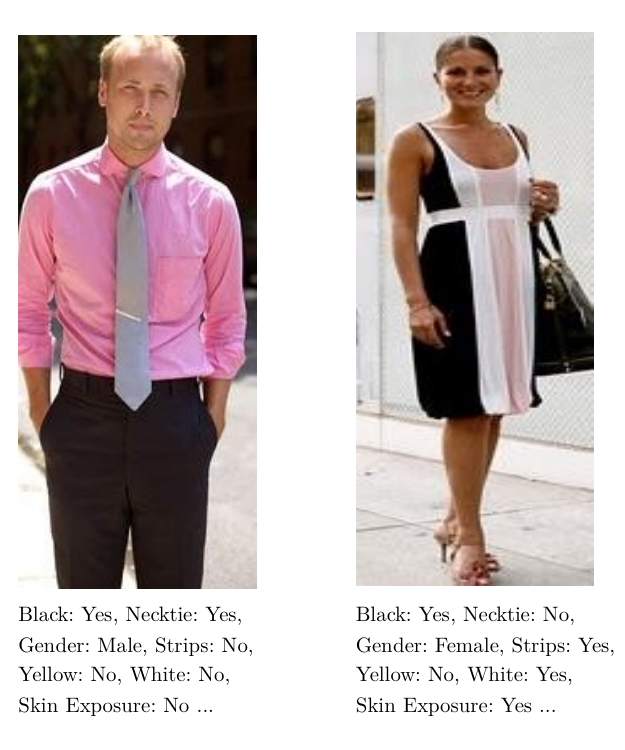}
		\caption {Illustration of binary semantic attributes. Examples from Clothing Attribute Dataset \cite{ChenDescribingClothing}. Yes/No indicates the existence/absence of the corresponding attribute.	\label{fig:semanticAttribute}}
	\end{center}
\end{figure}

Typically, discriminative learning approaches are used to learn semantic attributes (attributes that have names)\cite{facever_iccv2009, Farhadi09describingobjects,Lampert09learningto}. Figure~\ref{fig:semanticAttribute} shows two examples from the Clothing Attributes Dataset \cite{ChenDescribingClothing}, where both images have different attribute labels. Other types of attributes, such as data-driven ones, are learned in an unsupervised or weakly supervised manner \cite{ma:unsupervised}. Relative attributes are also introduced and learned through ranking methods (relative attributes have real values to describe the strength of the attribute presence) \cite{conf/iccv/ParikhG11, relativePartsAttributes}. However, most of the existing literature employs discriminative classifiers independently to predict the semantic attributes of low-level image features\cite{Farhadi09describingobjects, calibration6248021,Lampert09learningto}. Very few works model the relationship between object attributes, considering the fact that they may co-occur simultaneously in the same image\cite{jointLearningAttributes, dinesh-cvpr2014, learningShareVisualMulticlassDetection}.

Engineered low-level features like SIFT and HOG are used in combination with various loss-objective functions for attribute prediction purposes\cite{dinesh-cvpr2014, Farhadi09describingobjects,Lampert09learningto}. Improving the prediction results gives a good indication of the successful knowledge transfer of attributes between target tasks, for example, recognizing presently unseen classes through the transfer of attributes from another seen class\cite{JayaramanG14,Lampert09learningto }. In the work of \cite{Farhadi09describingobjects}, attribute models are learned to generalize across different categories by training a naive Bayes classifier on the  ground truth of semantic attributes. Then, they train linear SVM to learn non-semantic feature patterns and choose those which can be well predicted using the validation data attribute. The benefits of such generalization can be seen across different object categories, not just across instances within a category.

On the other hand, deep CNN demonstrates superior performance, dominating the top accuracy benchmarks in various vision problems\cite{NIPS2012_4824, veryDeepCNNs, DevilDeepCNN, deepCNNRankingMultilabels}. It has also been shown that CNN is able to generate robust and generic features\cite{cnnFeatureOffTheShelf, BengioRepresentation}. From a deep learning point of view, CNN learns image features from raw pixels through several convolutions, constructing a complicated, non-linear mapping between the input and the output. The lower convolution layers capture basic ordinary phrasing features (e.g., color blobs and edges), and the top layers are able to learn more complicated structure (e.g., car wheel)\cite{VisualizingCNN}. Subsequently, it is believed that such implementation of artificial Neural Networks mimics the visual cortex in the human brain \cite{NIPS2012_4824}.

Attribute prediction introduces two additional issues besides the typical object recognition challenges: image multi-labeling and correlation-based learning. Compared to single label classification, multi-labeling is more difficult. It is also more likely to appear in real word scenarios, at which point, an image would need to be tagged with a set of labels from a predefined pool \cite{Read:2009:CCM:1617459.1617477F, Tsoumakas07multi-labelclassification:}. For example, when a user types their query to search for an image, such as 'red velvet cake', the engine should retrieve consistent results for real cake images having a red velvet appearance. This is a difficult task from a computational perspective due to the huge hypothesis space of attributes (e.g., M attributes required $2^{M}$). This limits our ability to address the problem in its full form without transforming it into multiple single classification problems \cite{DBLP:journals/corr/WeiXHNDZY14}. In particular, correlations should be explored between all these singular classifiers to allow appropriate sharing of visual knowledge. Multi-task learning is an effective method for feature sharing, as well as competition among classifiers (the so-called 'tasks' in the term multi-task)\cite{dinesh-cvpr2014, learningToShareLatent}. If tasks are related and especially when one task lacks training data, it may receive visual knowledge from other, more fit tasks\cite{Caruana:1997:ML:262868.262872, icml2014c2_lic14, tree-guidedgroupMultitask, sharingFeaturesObjectsAttributes, multitaskFeatureLearning}. In the work of \cite{dinesh-cvpr2014}, they jointly learn groups of attribute models through multi-tasking using a typical logistic regression loss function.

Given the aforementioned issues, we propose an enhanced multi-task framework for an attribute prediction problem. We adapt deep CNN features as our feature representations to learn semantic attributes. Because the structure of the CNN network is huge, and thus requires powerful computation ability, we employ the following methods: First, if the number of attributes is small, we train multi-task CNN models together through MTL, where each CNN model is dedicated to learning one binary attribute. Second, if the number of attributes is relatively large, we fine-tune a CNN model separately on each attribute annotation to generate attribute-specific features, and then we apply our proposed MTL framework to jointly learn classifiers for predicting these binary attributes. The first approach is more or less applicable  depending on the available resources (CPU/GPU and Memory). The visual knowledge of different attribute classes can be shared with all CNN models/classifiers to boost the performance of each individual model. Among the existing methods in multi-tasking, the work in \cite{learningToShareLatent} proposes a flexible method for feature selection by introducing a latent task matrix, where all categories are selected to share only the related visual knowledge through this latent matrix, which can also learn localized features. Meanwhile, the work in \cite{dinesh-cvpr2014} interestingly utilizes the side information of semantic attribute relatedness. They used structured sparsity to encourage feature competition between groups and sharing within each of these groups. Unlike the work in \cite{learningToShareLatent}, we introduce the natural grouping information and maintain the decomposition method to obtain the sharable latent task matrix and thus flexible global sharing and competition between groups through learning localized features. Also, unlike the work of \cite{dinesh-cvpr2014}, we have no mutual exclusive pre-assumptions, such that groups may not overlap with each other in terms of attribute members. However, as the hand-crafted feature extraction methods can limit the performance, we exploit deep CNN models to generate features that better suit the attribute prediction case.

We test our method on popular benchmarks attribute datasets: Animals with Attributes (AwA)\cite{learningDetectAttributeTransfer} and the Clothing Attributes Dataset \cite{ChenDescribingClothing}. The results demonstrate the effectiveness of our method compared to standard methods. Because the Clothing dataset contains a small number of attributes, we successfully train our multi-task CNN model simultaneously. In addition because the AwA dataset contains a relatively large number of attributes, we first train each single CNN model on a target attribute. Then, we apply our multi-task framework on the generated features without instant back-propagation.

Our main contributions in this paper are summarized as follows: 1) We propose an enhanced deep CNN structure that allows different CNN models to share knowledge through multi-tasking; 2) We propose a new multi-task method; we naturally leverage the grouping information to encourage attributes in the same group to share feature statistics and discourage attributes in different groups to share knowledge. We relax any constraints on the groups, such as mutual exclusion, by decomposing the model parameters into a latent task matrix and a linear combination weight matrix. The latent task matrix can learn more localized feature, thus maintaining the ability to select some basic patterns through its configuration.

The remaining parts of our paper are summarized as follows: We first discuss the related work in Section \rom{2}. The proposed method for the Multi-task CNN model in addition to the details of our MTL framework are presented in Section \rom{3}. Experiments on two known attribute datasets and results are demonstrated in Section \rom{4}. Finally, we conclude the paper in Section \rom{5}.


 




\section{Related Work}
Because this work is mainly related to the topics of Semantic attributes, Multi-task learning and Deep CNN, we briefly review the most recent literature on these approaches including the following.

\subsection{Semantic Attributes}

\emph{Definition of Attribute:} a visual property that appears or disappears in an image. If this property can be expressed in human language, we call it a Semantic property. Different properties may describe different image features such as colors, patterns, and shapes \cite{Farhadi09describingobjects}. Some recent studies concentrate on how to link human-interaction applications through these mid-level attributes, where a consistent alignment should occur between human query expressions and the computer  interpretations of query attribute phrases.

\emph{Global vs. Local Attributes:} an attribute is global if it describes a holistic property in the image, e.g., 'middle aged' man. Usually, global attributes do not involve specific object parts or locations \cite{relativePartsAttributes, pandaPoseAttributeModeling,dinesh-cvpr2014}. Localized attributes are used to describe a part or several locations of the object, e.g. 'striped donkey'. Both types are not easy to infer, because if the classifier is only trained on high-level labels without spatial information like bounded boxes, the performance of the under-sampled classifiers may degrade. However, some work in \cite{dinesh-cvpr2014, learningToShareLatent} show that sharing visual knowledge can offset the effects of the lack of training samples.

\emph{Correlated Attributes:} If attributes are related and cooccur they are correlated.  In other words, some attributes will naturally imply others (e.g., 'green trees' and 'open sky' will imply 'natural scene'), so this configuration will impose some hierarchical relationship on these attribute classifiers. From another angle, attributes can be weaved from the same portion of the feature space and can be close to each other, e.g., 'black' and 'brown' attribute classifiers should be close to each other in the feature dimension space, belonging naturally to the same group, that is, the same color group \cite{dinesh-cvpr2014}. While most of the existing methods train independent classifiers to predict attributes \cite{Farhadi09describingobjects,Lampert09learningto,understandingObjectsAttributes}, typical statistical models, like naive Bayesian and structured SVM models, are used to address the problem. In \cite{Farhadi09describingobjects}, the authors employ a probabilistic generative model to classify attributes. In the work of \cite{Lampert09learningto}, objects are categorized based on discriminative attribute representations. Some works flow by modeling the relationships between classes with pre-assumptions of existing attribute correlations \cite{multitaskFeatureLearning}. Unlike this work, the decorrelation attribute method to resist the urge to share knowledge is proposed in \cite{dinesh-cvpr2014}, and they assume that attribute groups are mutually exclusive. Other work in \cite{relativePartsAttributes} proposes jointly learning several ranking objective functions for relative attribute prediction.

\emph{Attributes and Multi-labeling:} Image multi-labeling is simply learning to assign multiple labels to an image \cite{Tsoumakas07multi-labelclassification:, LEGO2-activeLearning}. If the problem is adapted as is, a challenge arises when the number of labels increases and the potential output label combinations become intractable \cite{Tsoumakas07multi-labelclassification:}. To mitigate this, a common transformation way is performed by splitting the problem into a set of single binary classifiers \cite{Read:2009:CCM:1617459.1617477}. Predicting co-occurring attributes can be seen as multi-label learning. On the other hand, most of the related works \cite{dinesh-cvpr2014,relativePartsAttributes} tend to apply multi-task learning to allow sharing or using some label relationship heuristics a priori \cite{labelRelationGraphs}. Another work applies ranking functions with deep CNN to rank label scores \cite{deepCNNRankingMultilabels}.

\subsection{Multi-task learning}
\emph{Why Mutli-task learning (MTL)?} MTL has recently been applied to computer vision problems, particularly when some tasks are under-sampled \cite{multitaskFeatureLearning, Caruana:1997:ML:262868.262872}. MTL is intended to impose knowledge sharing while solving multiple correlated tasks simultaneously. It has been demonstrated that this sharing can boost the performance of some or sometimes all of the tasks \cite{Caruana:1997:ML:262868.262872}.

\emph{Task and Feature Correlations:} Many strategies for sharing have been explored; the first one considers designing different approaches to discover the relationships between tasks \cite{Hariharan_largescale}, while the other considers an approach that aims to find some common feature structure shared by all tasks or mine the related features \cite{Rai_infinitepredictor}. Recent frameworks, like Max-margin \cite{Zhang_learningmultiple}, Bayesian \cite{icml2013_yang13a}, and their joint extension \cite{icml2014c2_lic14}, try to discover either or both task and feature correlations. While Max-margin is known by its discriminative power, Bayesian is more flexible and thus better suited to engage any a priori or performance inference \cite{icml2014c2_lic14}. In contrast to these studies, the work in \cite{Gong:2012:RMF:2339530.2339672} claims that as in typical cases, the dimension of the data is high; thus, the assumption that all the tasks should share a common set of features is not reasonable. They address such assumptions by simultaneously capturing the shared features among tasks and identifying outliers through introducing an outlier matrix \cite{Gong:2012:RMF:2339530.2339672}.
In other works, \cite{ICML2012Kumar_690, learningToShareLatent}, the authors further relax the constraint naturally by decomposing the model parameters into a shared latent task matrix and linear combination matrix; hence, all the tasks are encouraged to select what to share through this latent matrix, which can learn more localized features. However, among all these techniques, they rely on popular ways to perform such sharing through applying various regularizations on the model parameters, such as structure sparsity for feature selection and feature competition \cite{multitaskFeatureLearning}.

\subsection{Deep CNN}

\begin{figure*}
	\centering
	\begin{center}
		\includegraphics[width=1\textwidth]{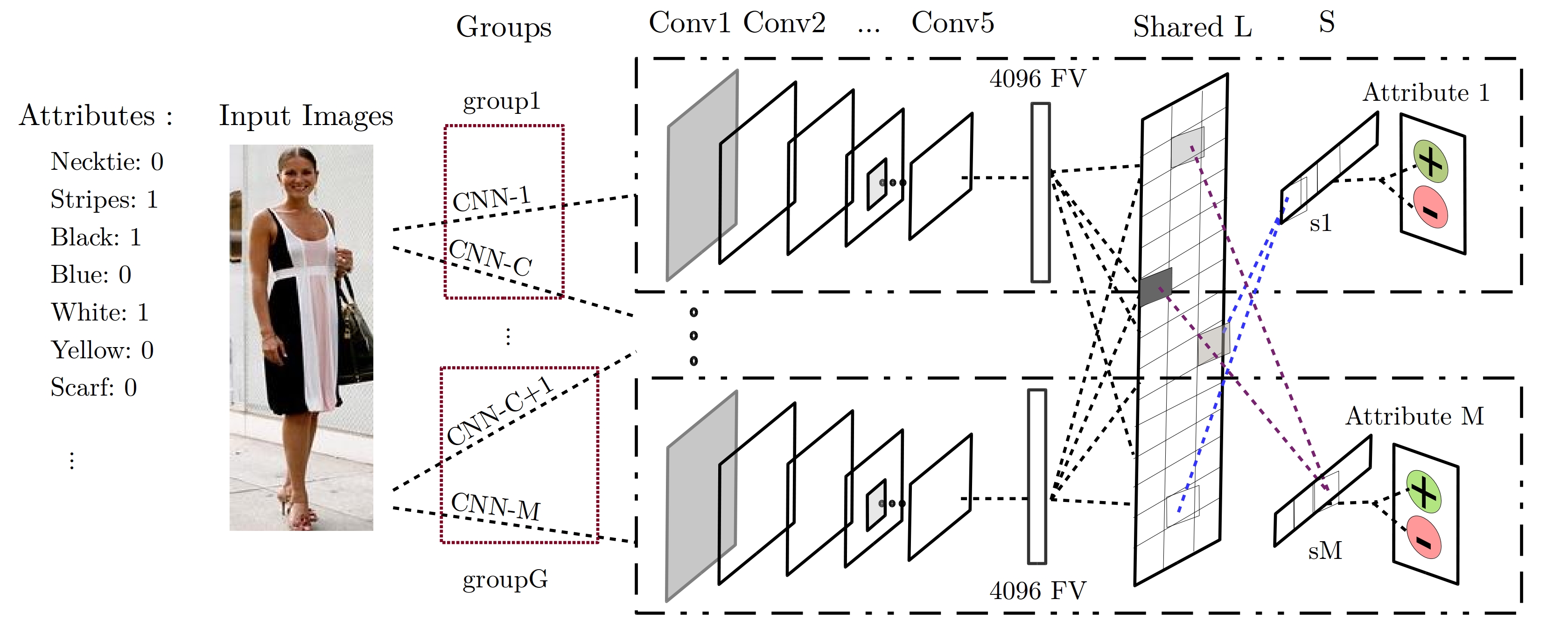}
	\end{center}
	\caption {Multi-task CNN models: the input image(in the left) with attribute labels information is fed into the model. Each CNN will predict one binary attribute. The shared layer $L$ together with the $S$ layer form a weight matrix of the last fully connected layer followed by a soft-max. The $L$ layer is a shared latent matrix between all CNN models. Each vector in $S$ is CNN-specific weight matrix layer. The soft-max and loss layers are replaced by our multi-task squared hinge loss. Group information about attribute relatedness is utilized during the training of the network.\label{fig:multitaskCNN}}
\end{figure*}

\emph{CNN for feature learning:} CNN was born in the Deep Learning (DL) era \cite{BengioRepresentation,NIPS2012_4824, DevilDeepCNN}; its goal is to model high-level abstractions of visual data by using multiple non-linear transformation architectures. Among the DL models, CNN shows extraordinary performance, specifically in image classification and object recognition applications \cite{NIPS2012_4824, DevilDeepCNN}. Two bothersome issues about training CNN are the number of training samples needed and the time that is required to fully train the network. This means that to have an effective CNN model, the training dataset and time should be large enough for the CNN to gain the ability to perform its task well \cite{NIPS2012_4824, veryDeepCNNs}. The learned features generated by the CNN are shown to be robust, generic, and more effective than hand-crafted features \cite{decafCNN, cnnFeatureOffTheShelf}. Fortunately, popular implementations of \cite{decafCNN, NIPS2012_4824} alongside the usage of pre-trained CNN models on the Imagenet dataset \cite{NIPS2012_4824} make it easier to fine-tune various CNN architectures for many vision datasets.

\emph{CNN between single and multi-labels:} using CNN for single label prediction is intensively studied \cite{NIPS2012_4824,BengioRepresentation }. There are many challenges that accompany multi-labeling, as previously discussed in the 'Attributes and Multi-labeling' section. Hence, training CNN directly is infeasible and impractical. However, one recent work proposes a work-around solution for the multi-label problem \cite{DBLP:journals/corr/WeiXHNDZY14}. In this work \cite{DBLP:journals/corr/WeiXHNDZY14}, a shared CNN is fed with an arbitrary number of object segment hypotheses (image batches), which are extracted or generated by some techniques, like the binarized normed gradients (BING) \cite{BingObj2014}. The final CNN output results for all of these hypotheses are aggregated by max-pooling to give the final format of the multi-label predictions. Unlike their approach, our proposed model holds the essence of tagging one image with multiple labels through multi-task CNN models, which are simultaneously trained through MTL to allow sharing of the visual knowledge. Another direction for multi-labeling is proposed by \cite{cnnFeatureOffTheShelf}, where the CNN is mainly used to generate off-the-shelf activation features; then, they apply SVM for later classifications. In our approach, when the number of attributes is large, we fine-tune many CNN models, each of which is dedicated to learning attribute-specific representations. These representations are used as off-the-shelf features for later stages in MTL, as we freeze their training while optimizing the multi-task loss function.

\emph{Convexity as first aids for CNN:} Some recent work \cite{cNNConvex, cNN-SVM} demonstrates that convex optimization can improve the performance of highly non-convex CNN models. The authors in \cite{cNNConvex} propose modifying the last two layers in the CNN network by making a linear combination of many sub-models and then replacing the original loss function by other ones from the convex optimization family. One of their findings is that hinge loss is one of the preferable convex functions that performs well during backpropagation. Another work \cite{cNN-SVM} confirms their finding that using the well-known SVM squared hinge loss does improve the final performance after training the CNN. By utilizing such experimental integration findings, we adopt a squared hinge loss framework to jointly optimize all classifier models while applying multi-tasking to naturally share visual knowledge between attribute groups.

In contrast to previous methods, our proposed approach is to train multi-task classifier models on deep features for attribute prediction and leverage a sharable latent task matrix that can be very informative for generating a full description of the input image in terms of attributes. Exploring the importance of such a latent matrix is a topic of future interest.

\section{Multi-task CNN Models}

In this section, we will explain the details of the proposed approach of the multi-task CNN model. Figure~\ref{fig:multitaskCNN} shows the overall structure of the proposed method, starting from raw images and ending with attribute predictions. Given a vocabulary of \emph{M} attributes, each CNN model will learn a binary nameable attribute. After the forward pass in all of the CNN models, the features generated from the last convolution layers will be fed into our joint MTL loss layer. To illustrate this more clearly, the weight parameter matrix learned in the loss layer will be decomposed into a latent task matrix and a combination matrix. The latent matrix can be seen as a shared layer between all the CNN models; in other words, the latent layer serves as a shared, fully-connected layer. Meanwhile, the combination matrix contains the specific information of each CNN model. It can also be seen as a specific fully-connected layer plugged above the shared latent fully-connected layer. After optimizing the joint loss layer and sharing the visual knowledge, each CNN model will take back its specific parameters through backpropagation in the backward pass. By presenting images that are annotated against several attributes, we iteratively train the whole structure until convergence. 

We adopt the popular network structure proposed by Krizhevsky \cite{NIPS2012_4824}, which consists of 5 convolutions, followed by 2 fully-connected layers and finally the softmax and the loss. In addition, some pooling, normalization, and ReLU are applied between some of these layers. Many works have studied and analyzed the nature of this structure and identified some important aspects. For example, the work in \cite{VisualizingCNN} shows that the parameters of the fully-connected layers occupy almost 70\% of the total network capacity, which consumes a great deal of effort while training the network. However, given the expense of trading between 'good-but-fast' and 'perfect-but-slow', the work in \cite{BengioRepresentation} shows that the performance will drop slightly when removing the fully-connected layers. Because our model requires more than one CNN model, we remove the last fully connected layers, as we substitute these layers with our own joint MTL objective loss, depending on the weight parameter matrix learned within.
 
 In the following subsections, we demonstrate the shared latent task matrix (which also can be seen as a shared layer in the multi-task CNN models approach). Then, we show how the feature sharing and competition is engaged. Next, we introduce our formulations, which we use to solve the attribute prediction problem. Finally, the total optimization procedure used to train the whole network of multi-task CNN models is described. In the remaining part of the paper, we will use the task/CNN model as an interchangeable meaning for classifier because in all cases we employ the same underlying MTL framework. The only difference is that in one approach, the attribute-specific feature learning is on-line, and the MTL joint cost function optimization changes will affect the bottom layers in all CNN models through back-propagation. Thus, any shared knowledge will also be back-propagated to the bottom layers. Meanwhile, in the other approach, we learn these attribute-specific features in isolation of optimizing the joint cost function, as training many on-line CNN models on a large number of attributes is impractical. 

\subsection{Sharing the Latent Task Matrix in MTL}
Given \emph{M} semantic attributes, the goal is to learn a binary linear classifier for each of them. Each classifier or task has model parameters, which are denoted by $w^{m}$ and are dedicated to predicting the corresponding attribute. $W$ is the total classifier weights matrix, which can also be considered a softmax weights matrix but stacked from all CNN softmax layers. Given $N$ training images, each of them has a label vector $Y$ of $M-dimension$, such that $Y_{m}^{i}$ is either \{1\} or \{-1\}, indicating whether a specific training image $i$ contains the corresponding $m$ attribute having a value of \{1\} or not \{-1\}. Suppose that the output from the last convolution layer in each CNN model forms our input feature vectors, such that each CNN model will generate an attribute-specific training pool. Thus, we will have $X^{N}_{M}$ training examples aggregated from all CNN models.

Our assumption is inspired from the work of \cite{learningToShareLatent}, where each classifier can be reconstructed from a number of shared latent tasks and a linear combination of these tasks. Through this decomposition, simultaneous CNN models can share similar visual patterns and perform flexible selection from the latent layer, which learns more localized features. We denote $L$ to be this latent task matrix, and $s^{m}$ is an attribute-specific linear combination column vector. In total, we have $S$ linear combination matrices for all attribute classifiers. 

Now, we want to split \emph{W} into two matrices \emph{L} and \emph{S}, as we assume that \emph{W} is a result of multiplying the shared \emph{L} latent matrix and the combination matrix \emph{S}, $W = LS$. To be more specific about each attribute classifier, the weight parameter vector can be formed by multiplying \emph{L} with the corresponding $s^{m}$ vector:
\begin{equation}
\begin{aligned}
\label{eg:decompose}
w^{m} = Ls^{m}
\end{aligned}
\end{equation}
where \emph{m} is the index of the m-{th} attribute, \emph{m = \{1,2,3 ... M\}}.

Given the CNN models, we aim to learn the matrix $ W $, which is formed by stacking the parameter matrices of the softmax layers of each CNN. The key idea behind our model is to decompose this weight matrix  $ W $ into two matrices $ L $ and $ S $, where the latent $ L $ matrix is the shared layer between all CNN models, $ S $ is a combination matrix, and each column corresponds to one CNN classification layer.

By this decomposition, each CNN can share visual patterns with other CNN models through the latent matrix $ L $, and all CNN models can collaborate together in the training stage to optimize this shared layer. Each CNN predicts whether the image contains the corresponding property. The benefit of learning the shared layer through multi-task is that each CNN can leverage the visual knowledge from learning other CNN models even if its training samples are not enough.

\subsection{Feature Sharing and Competition in MTL}

According to the semantic attribute grouping idea proposed in \cite{dinesh-cvpr2014}, group details are used as discriminative side information. It helps to promote which attribute classifiers are encouraged to share more visual knowledge, due to the group membership privileges. Meanwhile, different groups tend to compete with each other and share less knowledge. Table ~\ref{tab:AwAgroups} shows some group information examples from the Animal with Attribute (AwA) dataset\cite{learningDetectAttributeTransfer}. Because attributes are naturally grouped, we encode the grouping side information by encouraging attributes to share more if they belong to the same groups and compete with each other if they belong to different groups. Our attribute group information is shown in table \ref{clothingGroups}.

\begin{table}
		\caption{Examples of attribute groups from AwA dataset \cite{learningDetectAttributeTransfer}.	\label{tab:AwAgroups}} 
	\begin{center}
		\begin{tabular}{ | l | }
			\hline
			\emph{Texture:} \\
			patches, spots, stripes, furry, hairless, tough-skin \\\hline
			\emph{Shape:} \\ 
			big, small, bulbous, lean \\ \hline 
			\emph{Colors:} \\ 
			black, white, blue, brown, gray, orange, red, yellow\\\hline
			\emph{Character:} \\ 
			fierce, timid, smart, group, solitary, nest-spot, domestic \\ \hline
		\end{tabular}
	\end{center}
\end{table}

Suppose we have $M$ attributes and $G$ groups, where each group contains a variable number of attributes, for example, $g_{1}$ contains $[a_{1}, a_{2}, ... , a_{C}]$ as shown in the left side of figure~\ref{fig:multitaskCNN}, and each group can have a maximum of $M$ attributes. We have no restrictions on intra-group attributes. Even if two groups have the same attribute, the latent layer configuration mitigates the effect of the overlapped groups through the ability to learn more localized features. However, in our experiments, we rely on existing grouping information provided in the datasets, and obviously the groups are mutually exclusive (an attribute can be seen only in one group). Typically, solving the problem of overlapping groups requires some interior-point method, which is a type of second-order cone programming as discussed in \cite{Chen_smoothingproximal}, which is computationally expensive. Structured learning methods like group lasso \cite{treeGuidedLasso} are applied in many areas employing such grouping information. Knowing any a priori information about the statistical information of features will definitely aid the classifiers. Hence, in our MTL framework, we utilize rich information of groups and also adopt a flexible decomposition to learn different localized features through the latent matrix. We follow the work in \cite{dinesh-cvpr2014}, as they also applied such group information of attributes.

Regularizations are our critical calibration keys to balance feature sharing of intra-group attribute classifiers and feature competition between inter-group attribute classifiers. The idea is that when applying the $L_{1}$ norm as $\sum_{m=1}^{M} \|w\|_{1}$ \cite{singleTaskLasso}, it will consequently encourage the sparsity on both rows/features and columns/tasks of $W$. The effect of sparsity on the rows will generate a competition scenario between tasks; meanwhile, the sparsity effect on the columns will generate sparse vectors. Additionally, when applying the $L_{21}$ norm as $\sum_{d=1}^{D} \|w_{d}\|_{2}$ \cite{multitaskFeatureLearning}, where $D$ is the feature dimension space, in our case, because it is extracted from the previous layer, $D$ is $4096$. This can be seen as applying the $L_{1}$ norm on the zipped column-wise output of the $L_{21}$, which forces tasks to select only dimensions that are sharable by other tasks as a way to encourage feature sharing. As a middle solution \cite{dinesh-cvpr2014}, if the semantic group information is used when applying the $L_{21}$ norm, the competition can be applied on the groups; meanwhile, the sharing can be applied inside each group.

In our framework, we encourage intra-group feature sharing and inter-group feature competition through adapting the $L_{21}$ regularization term. Thus, we apply this on the vector set $s$ $\sum_{k=1}^{K}\sum_{g}^{G} \|s_{k}^{g}\|_{2}$ \cite{dinesh-cvpr2014, multitaskFeatureLearning}, where $K$ is the number of latent tasks (latent dimension space) and $G$ is the number of groups, where each group contains a certain number of attributes. Specifically, $s_{k}^{g}$ is a column vector corresponding to a specific attribute classification layer, and given a certain latent task dimension, $s$ will contain all the intra-group attribute vector sets. This will encourage attribute classifiers to share specific pieces of the latent dimension space, if they only belong to the same group. Meanwhile, different groups will compete with each other as each of them tries to learn a specific portion from the latent dimension space. Additionally,  the $L_{1}$ norm is applied on the latent matrix $\|L\|_{1}$ \cite{learningToShareLatent, singleTaskLasso}, to learn more localized visual patterns.

\subsection{Formulations of the Multi-task CNN model}
Given the above discussions about decomposing $W$ and by applying regularization alongside grouping information for better feature competition and sharing, we propose the following objective function:
\begin{equation}
\begin{aligned}
\label{fig:hingelossW}
\min_{L,S} \sum_{m=1}^{M} \sum_{i=1}^{N_{m}}& \frac{1}{2} [max(0, 1-Y_{m}^{i}(Ls^{m})^{T}X_{m}^{i})]^{2}\\
&+\mu \sum_{k=1}^{K}\sum_{g=1}^{G}\|s_{k}^{g}\|_{2} + \gamma\|L\|_{1} + \lambda\|L\|^{2}_{F}\\
\end{aligned}
\end{equation}
This is the typical squared hinge loss function, in addition to our extra regularizations. For the m-{th} attribute category, we denote its model parameter as $Ls^{m}$ and the corresponding training data is ${(X_{m}^{i},Y_{m}^{i})}_{i=1}^{N_{m}} \subset \mathbb{R}^{d} \times \{-1,+1\}(m=1,2,...,M)$, where $N_{m}$ is the number of training samples of the m-{th} attribute, and $K$ is the total latent task dimension space. In the second term and given a specific latent task dimension $k$, $s_{k}^{g}$ is a column vector that contains specific group attributes. The effect of this term is to continually elaborate on encouraging intra-group attributes to share feature dimensions. Thus, the columns/tasks in the combination matrix $S$ will share with one another only if they belong to the same group. Such competition between groups is appreciated; however, if there is some overlap between groups (they are not absolutely disjointed), some mitigation may help through the latent matrix $L$ configuration, which can learn more localized features. The $L_{1}$ norm is applied on the latent task matrix $L$ to enforce sparsity between hidden tasks. The last term is the Frobenius norm to avoid overfitting. Moreover, with such a configuration of the latent matrix $L$, an implicit feature grouping is promoted. Namely, the latent tasks will allow finding a subset of the input feature dimensions $D$, which are useful for related tasks, where their corresponding parameters in the linear combination matrix $S$ are nonzero.

Accordingly, every CNN is responsible for learning better input feature representations. Later in the testing, the input image will be fed into all CNN models to generate different input feature vectors; then the corresponding classifier weight vector will be applied to produce the attribute predications.  
 
The bottom layers in each CNN model are defined in the same way as the network structure proposed by \cite{NIPS2012_4824}. As shown in fig ~\ref{fig:multitaskCNN}, every block of CNN has several hidden layers, mainly 5 convolutions. We replace the last 2 fully connected layers, softmax and the loss by our proposed MTL squared maxing hinge loss layer. Nevertheless, when the number of attributes is large, we freeze the training of the bottom layers and optimize the multi-task loss function to predict attributes, using the outputs generated from the CNN models. 

\subsection{Optimization Steps}
Recall, that during the training procedure of $M$ CNN models, each of them is responsible for predicting a single attribute. Our goal is to impose visual knowledge sharing between all CNN models through optimizing the multi-task objective function. The optimized components of $W$ will serve as the last two fully connected layers. The $L$ component is a shared layer between all CNN models. The generalization ability of each single CNN is improved by leveraging the shared visual knowledge from other attribute classifiers. The burden of the Stochastic Gradient Descent (SGD) optimizer is only centralized in terms of training the bottom layers well if they are not freeze from training, so that each CNN can provide robust feature representation of images.

Solving the proposed cost function is non-trivial, because it is not jointly convex on either $L$ or $S$. The work in \cite{learningShareVisualMulticlassDetection} solves the non-convex regularized function using the block coordinate descent method. The function hence becomes a bi-form convex function. They employ Accelerated Proximal Gradient Descent (APG) to optimize both $L$ and $S$ in an alternating manner. Specifically, if $S$ is fixed, the function becomes convex over $L$, and optimizing it by APG can solve this state and handle the non-smooth effect of the $l_{1}$ norm. Likewise, if $L$ is fixed, the function becomes convex over $S$; in this form of the function and unlike \cite{learningShareVisualMulticlassDetection}, the mixed norm regularizations require re-representing the 2-norm into its dual form as discussed in \cite{dinesh-cvpr2014}. Smoothing Proximal Gradient Descent (SPGD) \cite{Chen_smoothingproximal, treeGuidedLasso, dinesh-cvpr2014} is applied to obtain the optimal solution of $S$. These optimizations are common in the literature of structured learning, where various regularizations may disturb convexity and smoothness properties of the functions. Algorithm $1$ illustrates the main steps that are applied to optimize equation 2.

Furthermore, when $L$ is fixed, the optimization problem is in terms of $S$ and is described as follows:
\begin{equation}
\begin{aligned}
\label{eg:hingelossW}
\min_{L,S} \sum_{m=1}^{M} \sum_{i=1}^{N_{m}}& \frac{1}{2} [max(0, 1-Y_{m}^{i}(Ls^{m})^{T}X_{m}^{i})]^{2}\\
&+\mu \sum_{k=1}^{K}\sum_{g=1}^{G}\|s_{k}^{g}\|_{2} \\
\end{aligned}
\end{equation}
\emph{Optimization by SPGD:} 
Chen et al. \cite{Chen_smoothingproximal} propose solving optimization problems that have a mixed-norm penalty over a priori grouping to achieve some form of structure sparsity. The idea behind this optimization is to introduce the smooth approximation of the objective loss function and solve this approximation function instead of optimizing the original objective function directly. Some work proposes solving non-overlapping groups, as is the case in \cite{dinesh-cvpr2014}. Others extend the solution to overlapping groups, as in \cite{Chen_smoothingproximal}. We closely follow the approach of approximating the objective function proposed in tree-guided group lasso \cite{treeGuidedLasso,treeGroupLassoeQTL}, which is basically built on the popular group-lasso penalty \cite{groupLassoOriginal}. We apply the step of squaring the mixed-norm term \cite{dinesh-cvpr2014}, which is originally suggested in \cite{groupLassoOriginal}. Squaring before optimization makes the regularizer positive, which generates a smooth monotonic mapping, preserving the same path of solutions but making the optimization easier. For further details on this approximation, refer to \cite{groupLassoOriginal}.
 
Now, after fixing $S$, the optimization problem is in terms of $L$ as follows:
\begin{equation}
\begin{aligned}
\label{eg:hingelossL}
\min_{L,S} \sum_{m=1}^{M} \sum_{i=1}^{N_{m}}& \frac{1}{2} [max(0, 1-Y_{m}^{i}(Ls^{m})^{T}X_{m}^{i})]^{2}\\
&+ \gamma\|L\|_{1} + \lambda\|L\|^{2}_{F}\\
\end{aligned} 
\end{equation}
\emph{Optimization by APG:} Accelerated Proximal Method updates the searching point from the last linear combination of two points in each iteration and thus converges faster \cite{acceleratedGM}. Furthermore, it also handles non-smooth convex functions using proximal operators. The idea is to rely on a shrinkage operator \cite{acceleratedGM, learningShareVisualMulticlassDetection} while updating the search point given the previous one and the gradient of the smooth part of the function (the non-regularized part). We adopt this method to optimize over $L$ because the proximity operator is straightforward, as the non-smooth $l_{1}$ norm has been studied extensively\cite{singleTaskLasso,groupLassoOriginal}. Meanwhile, in other general learning problems, the proximity operator cannot be computed explicitly, namely, the mixed-norm regularization term; hence, we adopt SPGD while optimizing $S$. We can optimize $L$ through SPGD by using approximations for both the gradient and the proximity operator; however, the APG has a relatively lower convergence rate.

\begin{table}
	\begin{center}
		\begin{tabular}{ |l }
\label{algr:optimizeeq2}
		 \textbf{\emph{Algorithm 1: Solving the Optimization Problem of Equation |2}} \\ \hline \\
		 \textbf{\emph{Input |}} \hspace{4pt}Generated features from CNN models : $X_{M}^{N}$ \\
\hspace{33pt} Attributes labels with values \{-1,1\} : $Y_{M}^{N}$ \\
		 \textbf{\emph{Output |}} Combination weight matrix $S$\\
\hspace{34pt} Latent tasks matrix $L$ \\
\hspace{34pt} Overall Model weight matrix $W$\\
		\textbf{\emph{Step 1 |}} Fix $L$ and optimize $S$ by SPGD \\
\hspace{34pt} Solving equation 3 until convergence\\
		\textbf{\emph{Step 2 |}} Get $S$ from Step 1, and optimize $L$ by APG\\
\hspace{34pt} Solving equation 4 until convergence\\
		\textbf{\emph{Step 3 |}} Repeat Step 1 and Step 2 \\
\hspace{34pt} Solving equation 2 until convergence\\
		\end{tabular}
	\end{center}
\end{table}

During a training epoch, the forward pass will generate the input for the multi-task loss layer from all the CNN models. After optimizing equation 2 using the proposed algorithm $1$, the output is the overall model weight matrix $W$, where each column in $W$ will be dedicated to its specific corresponding CNN model and is taken back in the backward pass alongside the gradients with respect to its input. $W$ is reconstructed using the optimal solutions of $L$ and $S$, where  knowledge sharing is already explored through MTL between all the CNN models via $L$.

\section{Experiments and Results}
\subsection{Datasets and Grouping}
We conduct our experiments on two datasets:\\
\emph{Clothing Attributes Dataset:} \\ This dataset is collected by the work in \cite{ChenDescribingClothing}; it contains 1856 images and 23 binary attributes, as well as 3 multi-class value attributes. The ground-truth is provided on image-level, and each image is annotated against all the attributes. We ignore the multi-class value attributes, because we are only interested in binary attributes. The purpose behind such clothing attributes is to provide better clothing recognition. We train Multi-task CNN models to predict the attributes in this dataset. Because no grouping information is suggested in this dataset, we follow the natural grouping sense proposed in other attribute datasets as in \cite{learningDetectAttributeTransfer}. In table~\ref{clothingGroups}, we show the details of our attribute grouping on the Clothing Attributes dataset.
\begin{table}
	\begin{center}
		\caption {Grouping information used in Clothing dataset\cite{ChenDescribingClothing}. \label{clothingGroups}}
		\begin{tabular}{ | l || c | }
			\hline
			\textbf{Group} &\textbf{Attributes}\\ \hline \hline
			Colors & black, blue, brown, cyan, gray, green, \\
			& many, red, purple, white, yellow \\ \hline
			Patterns & floral, graphics, plaid, solid, stripe, spot \\ \hline 
			Cloth-parts & necktie, scarf, placket, collar \\ \hline
			Appearance & skin-exposure, gender \\ \hline 
		\end{tabular}
	\end{center}
\end{table}

\emph{AwA Dataset:}\\ The Animals with Attributes dataset is collected in \cite{learningDetectAttributeTransfer}, the purpose of which is to apply transfer learning and zero-shot recognition \cite{AttributeBasedZeroRecogniton}. It consists of 30475 images of 50 animal classes. The class/attribute matrix is provided; hence, the annotation is on the animal's class level. It provides 85 binary attributes for each class. This dataset has 9 groups: colors, textures, shapes, activity, behavior, nutrition, character, habitat and body parts \cite{dinesh-cvpr2014}; table  ~\ref{tab:AwAgroups} shows some attributes in some of these groups.

\subsection{Attribute Prediction Accuracy}

We conduct several experiments on these two datasets. For the clothing dataset, we train multiple CNN models simultaneously. We calculate the accuracy of attribute predictions against the provided ground truth in the dataset. In table ~\ref{clothingREsults}, S-extract refers to a simple sitting where we directly use a pre-trained model of CNN \cite{imageNetCNN} for feature extraction, and then we train single SVM tasks for attribute prediction; meanwhile, in M-extract, we train our MTL framework on the same CNN extracted features. S-CNN refers to the single-task CNN, where we fine-tuned individual models of CNN to predict each attribute, and M-CNN refers to our MTL framework without encoding the group information \cite{learningToShareLatent}, and MG-CNN is our whole MTL framework with group encodings and wholly training CNN models with our framework together. CF refers to the combined features model with no pose baseline \cite{ChenDescribingClothing}, while CRF refers to the state-of-the-art method proposed by \cite{ChenDescribingClothing}. Our model outperforms the state-of-the-art results in \cite{ChenDescribingClothing}. We notice, though, that the overall improvement margin over the single CNN task models is relatively small compared to our results in AwA (see table ~\ref{AwAOverallREsults}). This is because the accuracy results are already quite high and thus hard to improve further.

\begin{table}[h!]
	\caption{The accuracy of attribute prediction before Sharing the \emph{L} layer, after sharing and previous methods on the clothing dataset \cite{describingClothingAttributes}. G1 refers to Color attributes, G2 refers to the Pattern group, G3 refers to Cloth-parts and G4 refers to the Appearance group. MG-CNN is our overall proposed framework.The higher, the better. For further details about several sitting and method names in this table, refer to Section \rom{4}-B.\label{clothingREsults}}
	\begin{center}
		\begin{tabular}{ |l||c|c|c|c|c|}
			\hline
			\textbf{Method} &\textbf{G1} &\textbf{G2} & \textbf{G3} &\textbf{G4} &\textbf{Total}\\ \hline \hline
			\textbf{S-extract} & 81.84 & 82.07 & 67.51 & 69.25 & 78.31 \\ \hline
			\textbf{M-extract} & 84.98 & 89.89 & 81.41 & 81.03 & 85.29 \\ \hline
			\textbf{S-CNN} & 90.50 & 92.90 & 87.00 & 89.57 & 90.43 \\ \hline
			\textbf{M-CNN} & 91.72 & 94.26 & 87.96 & 91.51 & 91.70 \\ \hline
			\textbf{MG-CNN} & 93.12 & 95.37 & 88.65 & 91.93 & \textbf{92.82} \\ \hline
			\textbf{CF\cite{compinedFeaturesNoPose}} & 81.00 & 82.08 & 77.63 & 78.50& 80.48 \\ \hline
			\textbf{CRF\cite{ChenDescribingClothing}} & 85.00 & 84.33 & 81.25 & 82.50& 83.95 \\ \hline
		\end{tabular}	
	\end{center}
\end{table}

\begin{figure*}
	 	 \centering
	\begin{center}
		\includegraphics[width=0.90\textwidth]{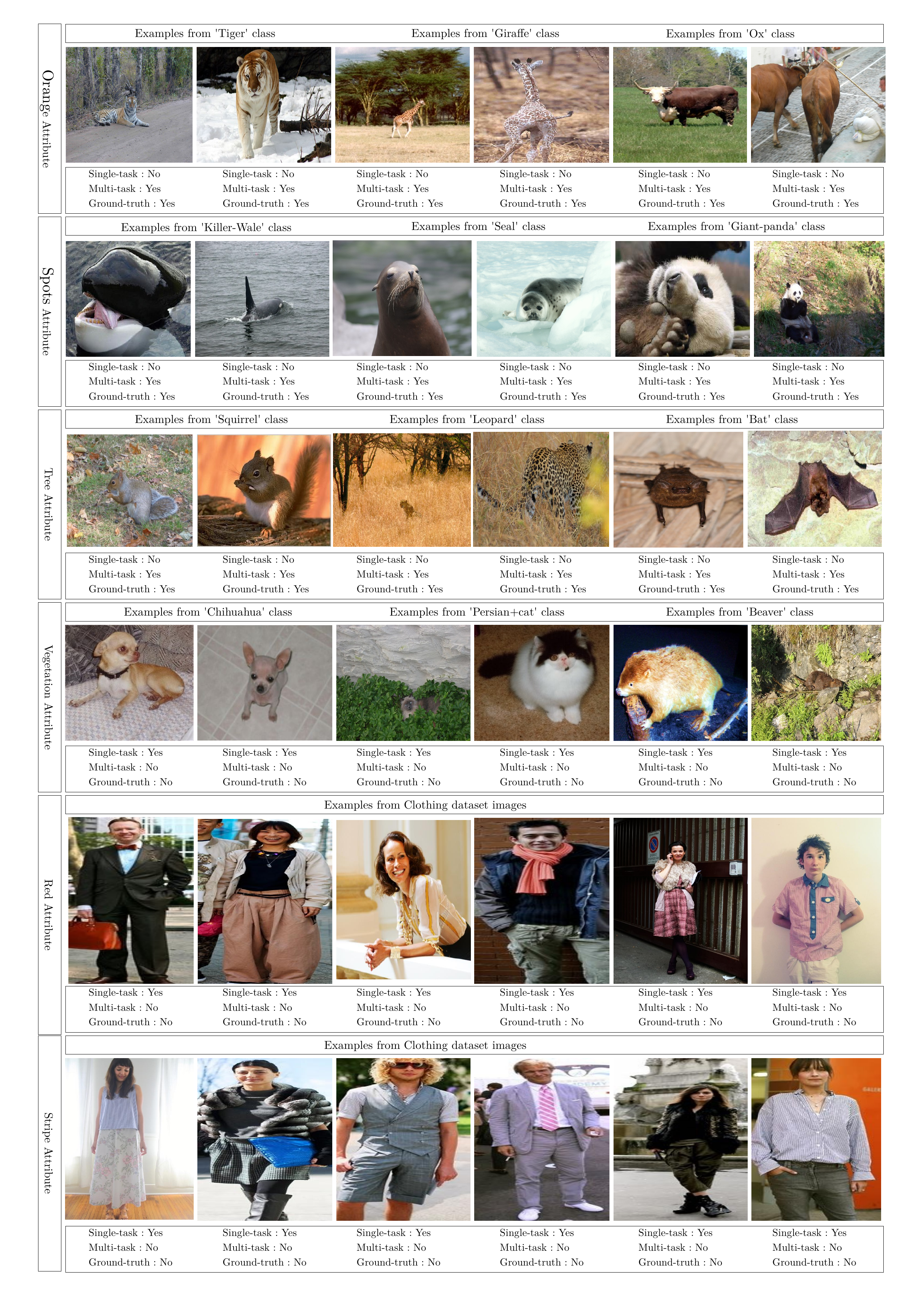}
	\end{center}
	\caption {Examples of misclassified samples from single-task classifiers results. The first 4 rows have samples from AwA dataset \cite{learningDetectAttributeTransfer}. The last 2 rows are samples from Clothing attributes dataset \cite{ChenDescribingClothing}. Yes/No indicates whether the attribute is presented or absent in the image. Multi-task classifiers are able to correctly classified these samples. \label{fifMiss}}
\end{figure*}

We conduct another experiment on the AwA dataset. We fine-tuned single CNN models separately on each attribute. Later, given the input images, we use these learned models to extract attribute-specific features. In other words, we freeze the training of the bottom layers in all CNN models and elaborate only in training our multi-task loss layer. This is due to the large number of attributes in the AwA dataset. We note that the fine-tuning stage will not add much practical difference and is a very time consuming process, perhaps due to the fact that AwA and Image-net datasets have an overlap of approximately 17 object categories; this has also been explored by another work \cite{understandingAttrsACCV}, in which they even train the CNN model to classify objects on the AwA dataset; however, they noticed that using the pre-trained CNN model on the Imagenet dataset directly or fine-tuning the model on AwA will in both cases give the same attribute prediction results. However, our MTL framework outperforms the single-tasks by a large margin; table ~\ref{AwAOverallREsults} shows the performance of our method compared with other standard methods, where the prediction accuracy is in terms of the mean average over all of the attributes. Compared with previous state-of-the-art results, which are approximately 75\% \cite{understandingAttrsACCV}, our trained MTL CNN models on attributes outperform it by a large margin. Additionally, table ~\ref{AwAResults} shows the accuracy results in terms of mean average precision over each group of attributes (group-level accuracy), before and after applying our multi-task framework. Initialization of the pre-trained model on Imagenet \cite{NIPS2012_4824} again is used throughout our experiments. Figure ~\ref{fifMiss} also shows a number of misclassified test samples from single-task classifiers, which our multi-task classifiers classified correctly.

\begin{table}
	\begin{center}
		\caption {Attribute detection scores of our Multi-task framework compared with other methods on AwA \cite{learningDetectAttributeTransfer}. The higher, the better. (mean average precision). \label{AwAOverallREsults}}
	
		\begin{tabular}{ | l || c |}
			\hline
			\textbf{Tasks} & \textbf{ Prediction Score} \\ \hline
			lasso \cite{singleTaskLasso} & 61.75 \\ \hline
			$l_{21}$ all-sharing \cite{multitaskFeatureLearning}& 60.21  \\ \hline
			$l_{2}$ regression loss & 66.87  \\ \hline
			decorrelated \cite{dinesh-cvpr2014}  & 64.80 \\ \hline
			category-trained CNN\cite{understandingAttrsACCV}  & 74.89 \\ \hline
			single CNN & 75.37 \\ \hline
			multi-task CNN (ours) & \textbf{81.19}\\ \hline
		\end{tabular}
	\end{center}
\end{table}

\begin{table}
	\begin{center}
		\caption {The group-level accuracy results of our Multi-task framework on AwA \cite{learningDetectAttributeTransfer}. The higher, the better.	\label{AwAResults}}
		\begin{tabular}{ | l || c | c | c |  }
			\hline
			\textbf{Groups} & \textbf{\# Attributes} &\textbf{Single CNN} & \textbf{Our Multi-task CNN} \\ \hline \hline
			Colors & 9 & 76.91 & 82.28 \\ \hline
			Texture & 6 & 76.16 & 82.44 \\ \hline
			Shape & 4 & 61.67 & 72.68 \\ \hline
			Body-Parts & 18 & 75.93 & 81.82 \\ \hline
			Activity & 10 & 82.22 & 85.4 \\ \hline
			Behavior & 5 & 72.78 & 74.96 \\ \hline
			Nutrition & 12 & 74.76 & 82.67 \\ \hline
			Habitat & 15 & 80.21 & 84.72\\ \hline
			Character & 7 & 62.01 & 70.52 \\ \hline \hline 
			\textbf{Total} & \textbf{85} & \textbf{75.37} &\textbf{81.19}  \\ \hline
		\end{tabular}
	\end{center}
	
\end{table}

\subsection{Implementation Details}
\emph{CNN model training:} We put each CNN model through 100 epochs, although most models converged in approximately 50 epochs. We initialize each CNN model with the pre-trained network on Imagenet \cite{NIPS2012_4824} and fine-tune it on the target attribute annotations. We normalize the input images into 256x256 and subtract the mean from them. To train our model on the Clothing dataset, we use a data augmentation strategy as in \cite{NIPS2012_4824}.\\\\
\emph{Multi-task Optimization:} We perform Singular Value Decomposition (SVD) on $W$ following the work in\cite{learningToShareLatent} to obtain an initialization for $L$; meanwhile, $S$ is randomly initialized. The initialization of $W$ was accomplished by stacking the last fully-connected layers from all pre-trained CNN models. The other model parameter values are either selected experimentally or following the typical heuristics and strategies proposed in \cite{NIPS2012_4824}. The latent tasks number is set to the maximum possible feature dimension, because the application of attribute prediction is very critical for any subtle fine-grained details; thus, any severe information loss caused via SVD can degrade the performance drastically. Hence, the number of latent tasks is set to 2048 in our experiments.

Additionally, we set the weight decay to 0.0005 in our CNN models; also, the momentum is set to 0.9, and the learning rate is initialized by 0.01 and reduced manually throughout training; we follow the same heuristic in \cite{imageNetCNN}. In our multi-task part (see equation 2), the latent task $\lambda\|L\|^{2}_{F}$ regularization parameter $\lambda$ is set to 0.4, and the other two parameters $\gamma$ and $\mu$ are best validated in each dataset experiments with held out unseen attribute data.

We conduct our experiments on two NVIDIA TK40 16GB GPU; the overall training time including the CNN part and MTL part of the training is approximately 1.5 days for the Clothing dataset (approximately 50 epochs for all 23 CNN models), and the testing time including feature extraction from all CNN models is approximately 50 minutes (sequential extraction from models one by one, not in parallel, where the time needed to extract features in each model is about 1.5 minutes/1000 images); if more attribute CNN models are added, the time will eventually increase. For the AwA dataset, we divide its training images into several sets (each set contains 3000 images, and we have 8 sets; 5 are used for training and 3 for testing). In total, the training time takes approximately 2 weeks (however, because we noticed that the major accuracy increase was mainly from training our MLT framework and not from CNN fine-tuning, we re-conducted the experiment and froze the bottom layers and depended on training the MTL layer, as we previously discussed; but in the second experiment, we saved a great deal of training time, as it only takes approximately 13 hours to completely train the last two layers on all CNN models within our MTL framework on one training set; on all remaining sets, it takes approximately 2.5 days to complete training.

\section{Conclusion}

In this paper, we introduce an enhanced multi-task learning method to better predict semantic binary attributes. We propose a multi-task CNN model to allow sharing of visual knowledge between tasks. We encode semantic group information in our MTL framework to encourage more sharing between attributes in the same group. We also propose decomposing the model parameters into a latent task matrix and a linear combination matrix. The latent task matrix can effectively learn localized feature patterns, and any under-sampled classifier will generalize better through leveraging this sharable latent layer. The importance of such a latent task matrix is a topic of future interest. Specifically, we would like to explore the potential of the latent task matrix decomposition to be informative enough to generate an efficient description of the input image in terms of either semantic or latent attributes. Our experiments on both attribute benchmark datasets show that our learned multi-task CNN classifiers easily outperform the previous single-task classifiers.


%



\section*{Acknowledgment}
The authors would like to thank NVIDIA Corporation for their donation of Tesla K40 GPUs used in this research at the Rapid-Rich Object Search Lab. This research is in part supported by Singapore Ministry of Education (MOE) Tier 2 ARC28/14, and Singapore Agency for Science, Technology and Research (A*STAR), Science and Engineering Research Council PSF1321202099. This research was carried out at both the Advanced Digital Sciences Center (ADSC), Illinois at Singapore Pt Ltd, Singapore, and at the Rapid-Rich Object Search (ROSE) Lab at the Nanyang Technological University, Singapore. This work is supported by the research grant for ADSC from A*STAR. The ROSE Lab is supported by the National Research Foundation, Singapore, under its Interactive \& Digital Media (IDM) Strategic Research Programme.



{\small
	\bibliographystyle{abbrv}
	\bibliography{egbib} 
}
\end{document}